# The Infrastructure Divide: Analyzing Community Stance on Data Center Development via Reddit

**Yoko Koyama[1], Gongfan Chen[2, *], Mansi Saxena[3]**

[1]Yoko Koyama, Lead Data Scientist, Carollo Engineers. Email: ykoyama@carollo.com

[2]Gongfan Chen, Assistant Professor, Albert School of Construction, Univ. of North Carolina at Charlotte, NC, USA 28223. Email: gchen12@charlotte.edu

[3]Mansi Saxena, Ph.D. Candidate, Department of Computer Science, North Carolina State University, Raleigh, NC, USA 27695. Email: msaxena4@ncsu.edu

* Corresponding Author, Gongfan Chen

## Abstract

The rapid expansion of AI-driven data center construction has outpaced public engagement: an estimated $64 billion in U.S. data center projects were blocked or delayed by local opposition between May 2024 and March 2025, yet the concerns driving this resistance remain poorly characterized. Prior engagement methods such as surveys and hearings are introduced late in the planning process, and no study has systematically mined social media to measure community stance at scale. To address this gap, this study presents an LLM-powered opinion-mining framework and applies it to 798 Reddit posts and 102,328 associated comments spanning a decade. An open-source large language model, guided by a Chain-of-Stance prompting framework and three domain-specific labeling rules, classified stance at 83% of the human inter-reviewer agreement ceiling. Opposition dominated discourse volume rather than disapproval: against posts outnumbered favor posts 2.6:1 and generated 8.5 times more comments, and this gap widened significantly over 86 weeks. Topic modeling of opposition arguments resolved into four reproducible clusters, resource consumption and scarcity (42.8%), contested siting processes and

secrecy (34.2%), political opposition to data centers (13%), and political and corporate accountability (8%). Engagement analysis revealed that posts with emotional or provocative political framing attracted the most engagement, while key technical concerns (e.g., water, power) were dispersed across lower-engagement comment threads. These findings give local leaders, developers, and policy makers a quantified, continuously updatable basis for prioritizing concerns and policy levers, supporting a shift from opposition to negotiated, verifiable partnership for data center development.

## Introduction

The rapid emergence of modern artificial intelligence (AI) has triggered a new wave of data center construction worldwide. Driven by the growing demand for high-performance computing, specialized hardware accelerators, and scalable cloud infrastructure, data centers have become critical infrastructures for supporting large-scale data storage, networking, processing, and AI computation (Avelar et al. 2023; Kong et al. 2024; Lei et al. 2025). Despite this rapid expansion, the supply of data center capacity continues to lag behind market demand. Coldwell Banker Richard Ellis (CBRE Research 2025) reported that North American data center vacancy rates fell below 3% in 2025, a historic low, as over 90% of new capacity was pre-leased before becoming operational. Market trends further highlight this imbalance, as McKinsey (2024) reported that prices charged by colocation providers in the United States increased by an average of 35% between 2020 and 2023, while much of the capacity expected to come online within the next two to three years had already been leased in advance. In response, governments, technology companies, and private industries are investing heavily in hyperscale and AI-oriented data centers to meet escalating computational demands and strengthen competitiveness in the digital economy (McKinsey 2025; The White House 2025).

However, this rapid growth also creates major challenges related to electricity consumption, cooling demand, grid capacity, water use, sustainability, and environmental impacts. For example, Goldman Sachs Research (2026) estimates that global data center demand will grow by about 17% per year, reaching nearly 95 GW by 2027, while total power capacity could exceed 90 GW by 2030, surpassing California's current electricity consumption. Many data centers also require substantial water for cooling and space conditioning, raising concerns about water use and local environmental impacts (Lei et al. 2025). In addition, carbon emissions from global data centers have increased significantly since 2018 and are projected to account for 3.2% of global emissions in 2025 and 14% by 2040 (Zhu et al. 2023). Due to these concerns, communities that host or are targeted for data center development have shown varying levels of opposition. Between May 2024 and March 2025, an estimated $64 billion in data center projects were blocked or delayed due to local resistance (Data Center Watch 2025). This highlights the need to better understand public opinions toward data centers, including the key concerns, influencing factors, and geographic or community scales at which opposition emerges. Such understanding can help local leaders identify what can be negotiated to shift the conversation from opposition to partnership, ensuring that data center development delivers tangible local benefits and supports mutually beneficial outcomes.

Social media data are increasingly used to crowdsource near-real-time information and public opinions because they capture digital traces of content sharing and public discussion (Chen et al. 2020). Researchers have used platforms such as X (formerly Twitter), LinkedIn, and Reddit to study different aspects of the built environment. For example, Twitter data and natural language processing (NLP) techniques have been used to examine public responses to disasters (Banerjee et al. 2026; Chen and Ji 2021; Yuan and Liu 2020). LinkedIn data have been used to study gender disparity and self-presentation among AEC industry leaders (Mansouri and Erfani 2024), gender differences in construction career paths (Hickey et al. 2022), and the gender likeability-competence dilemma (Erfani et al. 2023). Reddit data and graph-based analysis have also been used to examine construction safety issues, such as heat stress (Chen et al. 2026; D'Souza et al. 2026). Together, these studies show that social media provides rich, timely, and publicly available data that can be mined through NLP and semantic analysis. However, public opinions toward data centers on social media remain underexplored, especially in various online public discourse where concerns and expectations often emerge before they appear in formal planning processes. Social media offers a valuable lens for capturing how different stakeholders frame data centers, what trade-offs they emphasize, and how local debates evolve across regions and time. Analyzing these discussions can help identify recurring themes, misinformation, trust gaps, and community priorities, providing practical insights for project planning, public engagement, and policy design.

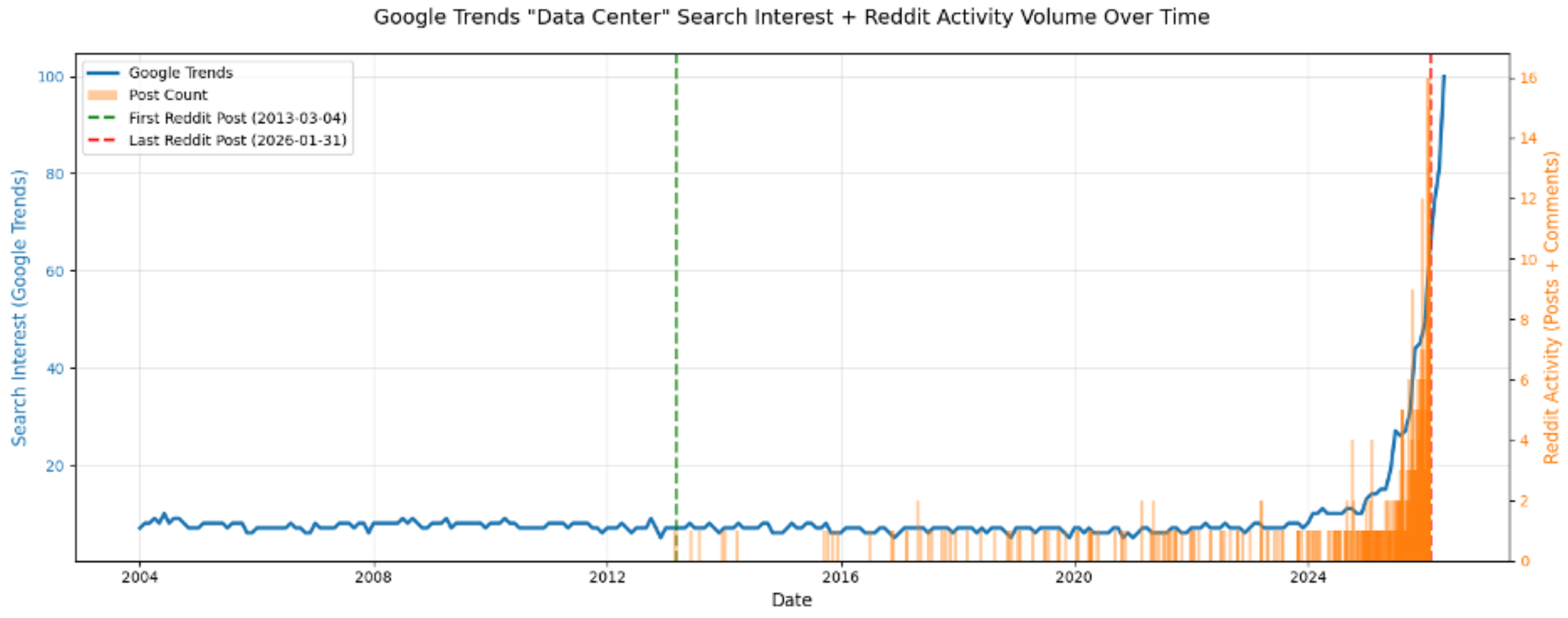

**Figure 1. Time series of google trends search term, "data center", and Reddit post volume**

To address this gap, this study mines Reddit data to examine public stances toward data center development and generate insights for decision-makers involved in project planning and community engagement, as well as policy makers at state or federal level. To assess whether Reddit data can serve as a proxy for broader online discourse, we compared the weekly volume of collected Reddit posts with Google Trends search interest for the term “data center.” Reddit discussion volume increased sharply beginning in 2025, closely aligning with the concurrent surge in Google search interest (Figure 1). This similarity suggests that Reddit discussions reflect broader public attention to the topic.

In this study, we collected 798 Reddit posts and 102,328 associated comments and developed an LLM-powered opinion-mining framework to examine public attitudes toward data center development. The analysis included stance detection, time-series analysis, engagement analysis, and topic modeling. Specifically, this study is guided by the following research questions:

(1) What are the overall public stances toward data center projects?

(2) How do public stances and discourse topics change over time?

(3) What factors shape supportive, opposing, or neutral stances?

The results of this study will provide leading indicators to inform policy considerations for data center infrastructure development, helping policymakers balance the interests of local communities and private-sector developers while responding to the growing global demand for computing capacity driven by AI.

The remainder of this paper is organized as follows. The Literature Review section reviews prior studies on community opinions related to data centers and key techniques for social media mining. The Methodology section describes data collection, data cleaning, and NLP- and LLM-based methods for social media analysis. The Results section presents public stances, major topics, and topic changes and engagements over time. The Discussion section highlights key insights, implications for decision-makers’ planning strategies, and study limitations. Finally, the Conclusion section summarizes the main findings and contributions.

## Literature Review

**Public Opinion on Data Centers**

Research on public opinion toward data centers remains limited. Existing scholarship has historically prioritized technical concerns, especially energy consumption, while giving less sustained attention to the social and political dimensions of this infrastructure (Terenius et al. 2023). As a result, the field still lacks a strong empirical foundation for understanding how communities perceive, value, or oppose data center development. This gap is especially important because data centers increasingly raise social and political questions that require dedicated study, rather than being treated as secondary to technical performance or environmental efficiency (Terenius et al. 2023).

In practice, public sentiment or stance has often been addressed through traditional polling and public hearings, but these forms of engagement are frequently introduced late in the development process rather than during early planning stages (Wu et al. 2025). Emerging research has begun to apply established public opinion methods, including representative surveys and qualitative interviews. For example, Pew Research Center (2026) found that 39% of Americans believe data centers negatively affect the environment, compared with 4% who see positive effects; 38% say data centers increase home energy costs, compared with 6% who report positive effects; and 30% believe they reduce quality of life for nearby residents, compared with 6% who report positive effects. In contrast, Airedale by Modine (2025) reported that 45% of respondents believed data centers would have a positive impact on their community, while 19% anticipated negative consequences. Together, these findings suggest that public opinion is not defined by simple rejection, but by conditional acceptance shaped by perceived costs, benefits, and local context.

Multi-method studies using interviews, focus groups, and site visits further show that communities often view data centers as necessary infrastructure for modern digital life, while still questioning whether local benefits justify local burdens. For example, residents may recognize the importance of data centers but remain skeptical of their limited employment benefits, which are often perceived as modest compared with other industrial developments (Sovacool et al. 2022). Research on local resistance has also identified multiple dimensions of public concern. Rone (2024) for instance, analyzed opposition in the Netherlands through qualitative content analysis of 111 regional newspaper articles and interviews with politicians and activists, identifying nine major topics of concern. Recent qualitative work in the United States similarly shows that residents often describe surprise at the speed of local transformation and express hopelessness about preserving their expected quality of life (Ngata et al. 2025). Related research on data-sharing infrastructure suggests that backlash is more likely when communities feel poorly informed, distrust privacy protections, or learn too late about commercial involvement, highlighting the importance of trust, transparency, and early communication in shaping public acceptance (Nwebonyi et al. 2022).

Recent scholarship therefore emphasizes the need for earlier, more inclusive, and more sustained engagement with affected communities. Rather than focusing narrowly on facility siting, researchers increasingly call for attention to service workers, activists, residents, and communities affected by energy supply, extraction, or remote sitting decisions (Edwards et al. 2025). Methodologically, scholars advocate for ethnographic and community-based approaches that can capture lived experience, shifting expectations, and forms of exclusion that short surveys may miss (Edwards et al. 2025). At the same time, newer data-augmented methods, including social media analysis and AI-assisted theme extraction, are beginning to expand how public discussion can be studied (Wang et al. 2025; Wu et al. 2025). For example, Wu et al. (2025) proposed using LLM agents as proxies for community opinion on data center projects and found that agent responses varied significantly by model and regional context while showing strong topical alignment with real-world survey data. However, this simulation-based approaches still face important limitations, including missing demographic context and challenges of verification (Ngata et al. 2025; Wang et al. 2025).

**LLM-Assisted Methodologies for Understanding Public Opinion, Sentiment, and Stance**

Transformer-based models and Large Language Models (LLMs) have transformed social media data mining, moving beyond generic text classification to capture nuanced public discourses around energy and physical infrastructure (Cruickshank and Ng 2026; Gambini et al. 2024). Methodologically, prior research relies on three core strategies: combining transformer-based sentiment analysis with topic modeling (Kastrati et al. 2023), fine-tuning LLMs on domain-specific datasets (Kim et al. 2021; Kwon et al. 2024), and deploying zero- or few-shot stance detection pipelines (Gambini et al. 2024; Steel and Ruths 2024). These methods are invaluable because they extract geographic, policy, and framing variations at a scale unachievable through traditional survey methods. For instance, social media mining has successfully mapped energy price anxiety (Kastrati et al. 2023), isolated partisan framing from core technical attitudes toward nuclear energy (Kwon et al. 2024) and demonstrated how state-level net-metering policies drive favorable solar energy sentiment (Kim et al. 2021).

Crucially, recent work shifts focus from overall sentiment (positive/negative tone) to target-specific stance detection (support, opposition, or neutrality toward a specific subject). To resolve accuracy inconsistencies in zero-shot LLM classification (Cruickshank and Ng 2026), scholars increasingly incorporate step-by-step reasoning frameworks. Methods such as Chain of Stance (Ma et al. 2024) and background-knowledge prompting (Taranukhin et al. 2024) significantly outperform baseline and fully supervised models by forcing LLMs to analyze contextual context before declaring a stance. Furthermore, expanding the unit of analysis from isolated posts to user posting histories (Gambini et al. 2024) or thread-level multi-target discussions (Steel and Ruths 2024) enables social media data mining to reliably mirror established public opinion polling. Moreover, topic modeling is an unsupervised machine learning technique used to uncover latent themes within large collections of textual data (Chowdhury and Zhu 2023). This approach identifies clusters of frequently co-occurring words, forming topics that reflect the underlying structure of the data (Vayansky and Kumar 2020; Wang and Taylor 2019). Transformer-based approaches such as BERTopic have demonstrated the potential to capture semantic meaning and contextual relationships among words more effectively, thereby producing more coherent and interpretable topics (Chen et al. 2026; Grootendorst 2022). Collectively, these studies demonstrate the potential of LLM-assisted methods for extracting nuanced public attitudes from social media.

## Methodology

### Overview

The overarching aim of this study was to characterize and understand public stance toward data center development using authentic, community-generated social media text data. The design of the study revolves around answering the following research questions (RQs):

- RQ i— What are the overall public stances toward data center projects? Specifically, we ask a) what is the stance distribution, and b) how popular are the different stances?
- RQ ii— How do public stances and discourse topics change over time?
- RQ iii— What factors shape supportive, opposing, or neutral stances and which topics drive engagement?

As shown in Figure 2, the methodologies to answer these research questions consisted of four steps: (1) Collection of Reddit posts and comments through keyword search, and data filtration through keywords exclusion, length of contents and engagement metrics; (2) Deploying open-source LLMs for stance classification (Favor, Against, Neutral), human-in-the-loop quality-control (QC) label generation, and empirical output validation; (3) Stance-stratified analyses of community engagement and temporal trends; and (4) Identification of argument clusters, keywords' temporal trends via topic modeling. Step (2) addresses RQ i.a (what are the public stances toward data center development and what is the stance distribution). Step (3) addresses RQ i.b (how popular are the different stances) and RQ ii (How do public stances and associated opinions change over time). Finally, step (4) addresses RQ iii (What factors shape supportive, opposing, or neutral stances).

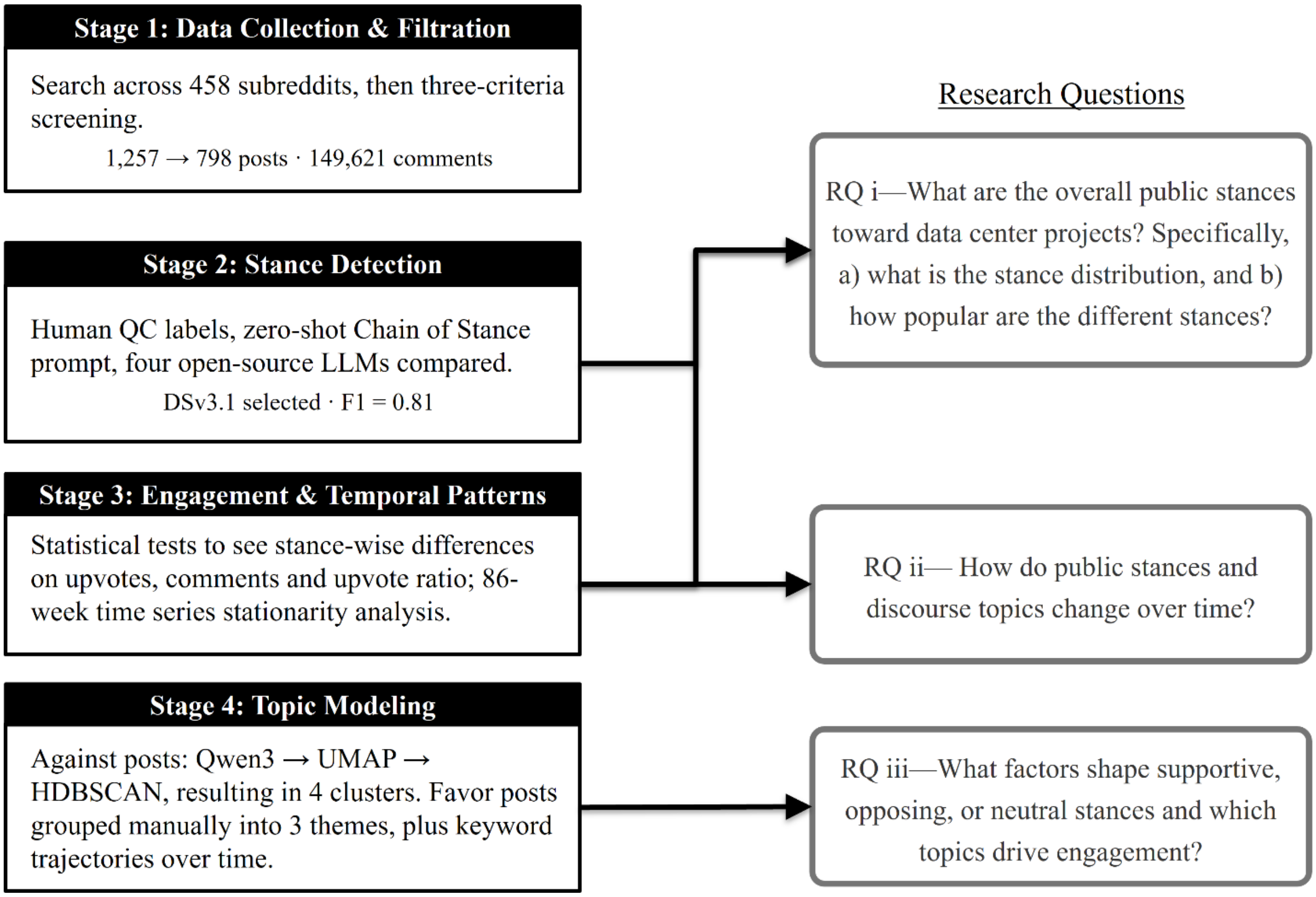

**Figure 2. Overview of four-step analytical pipeline and its connection to the three RQs.**

## Data Collection & Filtration

We retrieved Reddit posts and their associated comments through the official Reddit API using a two-layer keyword strategy (Text S1, SI). First, a platform-wide search of six seed queries covering American and British spellings and industry synonyms (e.g., "data center," "data centre," "server farms") returned 966 unique posts across 455 subreddits. Second, we converted the same seed terms into candidate subreddit names and harvested posts from the four communities that existed, adding 291 posts not captured by the first layer. The combined corpus comprised 1,257 posts and 149,621 comments spanning more than a decade of discussion (Table 1).

**Table 1. Summary of collected data statistics**

| Parameter | Value |
|---|---|
| **Total number of posts collected based on keyword search** | 1,257 |
| **Total number of comments collected based on keyword search** | 149,621 |
| **Number of posts after data filtration** | 795 |
| **Number of associated comments after data filtration** | 102,328 |

We then screened posts for relevance to data center development using three rules applied to the combined title and body text: removal of off-topic gaming content, removal of short posts with low engagement, and removal of posts with no comments (Text S1, SI). After we removed duplicate permalinks and posts linking to YouTube, filtration retained 64% of posts and 68% of comments: the final dataset comprised 798 posts and 102,328 comments (Table 1). We removed URLs and other non-linguistic artifacts from all text prior to analysis.

**Stance Detection**

Stance is defined as the expression of the speaker's standpoint and judgment toward a given proposition. Stance detection plays a major role in analytical studies measuring public opinion on social media, particularly on political and social issues. Stance detection has evolved from feature-engineered classical machine learning (ML) and deep learning to transformer-based and now LLM-based approaches, with each generation improving handling of implicit, context-dependent, and target-specific stance (Pangtey et al. 2026). In this study, we perform a unimodal (text-data-only), target-specific (focused on data center development) stance detection task using the open-source LLMs. The key methodology involves prompt engineering based on the *Chain-of-Stance* framework (Ma et al. 2024).

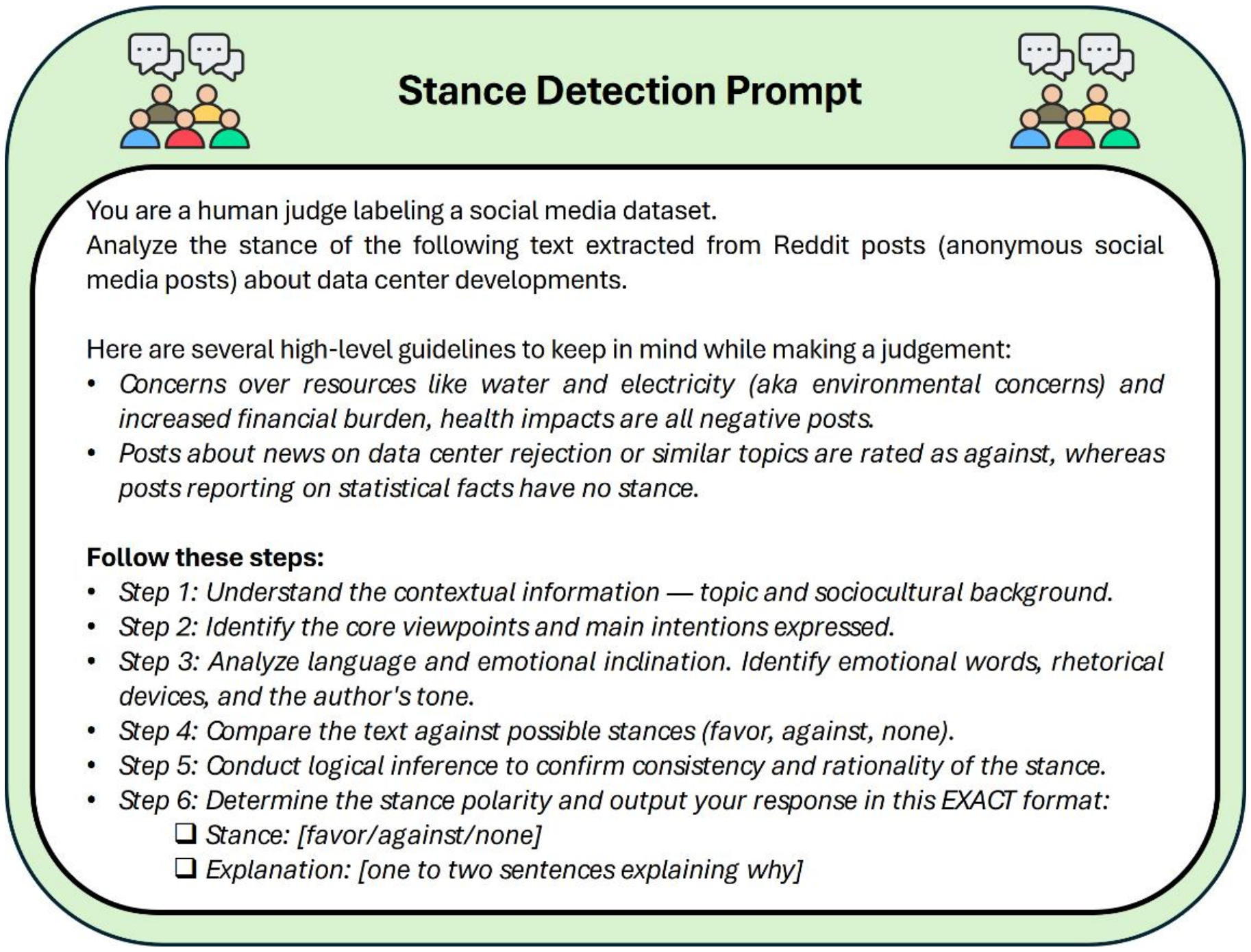

**Figure 3. Stance detection prompt used in this study**

Before initiating LLM-based labeling, two human annotators with professional engineering backgrounds—one from industry and one from academia—jointly reviewed an initial sample of social media posts through iterative reconciliation and reached a consensus on the criteria for labeling each post as favorable, neutral, or against. Rule 1: posts expressing concerns over resources such as water and electricity (i.e., environmental concerns), increased financial burden, or health impacts are labeled against. Rule 2: posts reporting news of data center rejections or similar events are labeled against, whereas posts reporting purely statistical facts carry no stance. Rule 3: when a benefit of data center development is mentioned, or push back on opposition, the post is labeled favor. Based on these rules, the prompt instructs the LLM to act as a human judge and follow a six-step reasoning process—from contextual interpretation to logical verification—to assign a stance and provide a one- to two-sentence explanation in a fixed output format. The following prompt was applied to all candidate LLM models to generate stances and enable direct comparison, as shown in Figure 3.

Four open-source LLMs were selected as candidate stance classifiers: (1) DeepSeek v3.1 (DSv3.1), (2) DeepSeek R1 (DSR1), (3) Llama 4, and (4) Llama 3. These models are selected based on performance of open-source LLMs benchmarked in a recent study that evaluated over twenty LLMs for sentiment classification and its public availability (Nasution et al. 2025). Further, we also compared a mixture-of-experts (MoE) model against dense model within the same model family. MoE model architecture is designed to provide more capability than a dense model given the same amount of computational hardware resource (Cai et al. 2024). To evaluate LLM-based automated stance detection, two human experts (same two who developed rules for stance classification) manually labeled the data, serving as the ground truth for assessing model performance. Human annotators reviewed and labeled a subset of 300 posts. Following Cochran (1977), this sample size was selected to achieve a 95% confidence level based on the size of the collected dataset. Then, each model classified this subset, and its output was compared with the labels of each human reviewer using weighted precision, recall, F1-score, and Cohen's $\kappa$. Inter-reviewer agreement on the shared double-annotated subset (Cohen's $\kappa = 0.786$, substantial agreement) was treated as the practical performance ceiling for the LLM classifiers, given that even trained human reviewers do not achieve perfect alignment. The top-performing model was used to classify the full corpus of posts. We conducted additional analysis detailed in supporting information (SI) on model error types for the top-performing model and other models (Text S2, SI). Posts and their associated comments were subsequently grouped by the assigned stance for downstream analyses.

## Stance Engagement and Temporal Pattern Analysis

In this study, user engagement with competing positions (RQ i.b) is defined by three post-level metrics: upvote score, number of comments, and upvote ratio. These data were directly available from the Reddit scraping results. A few statistical tests were used to determine whether the level of user engagement was higher in one stance group than another. Because engagement distributions differed markedly in shape and variance across stance groups (e.g., the standard deviation of comment counts ranged from 131 in the favor group to 2,681 in the neutral group), pairwise differences were tested with the Brunner–Munzel test. The test evaluated whether a randomly selected post from one group is more likely to show higher engagement than a randomly selected post from another.

To examine how public stance changed over time (RQ ii), we aggregated posts and comments into weekly counts for each stance group across 86 weeks. We also calculated the weekly differences between the against and favor groups (against count − favor count) for both posts and comments. We then tested each time series for stationarity using the Augmented Dickey–Fuller (ADF) and Kwiatkowski–Phillips–Schmidt–Shin (KPSS) tests, using a p-value of 0.05 as threshold. These tests have opposite null hypotheses and were therefore used together to reduce the risk of misclassification. A time series was considered nonstationary when the ADF test failed to reject the presence of a unit root and the KPSS test rejected stationarity.

## Topic Modeling

### *BERTopic Modeling*

To identify the reasons underlying each stance toward data center development (RQ iii), we applied BERTopic, an unsupervised framework that identifies latent themes in large text collections. BERTopic involves four steps: (1) generating contextual document embeddings with a pretrained language model; (2) reducing embedding dimensionality with Uniform Manifold Approximation and Projection (UMAP; McInnes et al. 2020); (3) clustering semantically similar documents with Hierarchical Density-Based Spatial Clustering of Applications with Noise (HDBSCAN; Malzer and Baum 2020); and (4) representing each topic with class-based term frequency–inverse document frequency (c-TF-IDF; Grootendorst 2022).

Rather than modeling the original posts, which contained links, nonsensical text, and code fragments, we used the stance explanation that DeepSeek v3.1 generated for each post as input. These explanations focus on the reasons behind a stance and provide cleaner text for topic modeling. We embedded each explanation with Qwen3-Embedding-8B (Zhang et al. 2025), an embedding model selected from a ranking framework (Muennighoff et al. 2023), reduced the embeddings to five dimensions with UMAP, and clustered them with HDBSCAN. We then applied c-TF-IDF to extract the most representative terms in each cluster, and DeepSeek v3.1 reviewed each cluster's terms and posts to generate a concise summary of its central theme.

We tuned HDBSCAN parameters by grid search, selecting the pair that maximized the Density-Based Clustering Validation index (DBCV; Moulavi et al. 2014), and assessed stability across 10 repeated runs (Text S3, Supplementary Materials (SMs)). The against-stance clusters were stable: all 10 runs produced four stable clusters with five outliers (Text S3, SM).

*Manual topic modeling*

Because the dataset of posts labeled *favor* was small, the authors identified themes manually rather than applying automated topic modeling. They grouped posts by their central arguments and assigned a theme to each group. Posts that were irrelevant or did not fit a coherent theme were excluded.

*Engagement-based topic modeling*

To identify the phrases associated with the highest engagement in posts opposing data center development, we used c-TF-IDF to extract 61 bigrams and trigrams from the titles and bodies of posts classified as *against*, after cleaning the text and removing seed words and stop words. We then ranked these phrases separately by post and comment engagement. For each phrase, post-level engagement was the sum of upvote scores and comment counts across posts containing it. Comment-level engagement was the sum of upvote scores across comments containing the phrase. We identified the top five phrases at each level, based on these respective engagement metrics.

*Temporal topic modeling*

Lastly, to examine how discussion topics evolved over time, we extracted and ranked bigrams and trigrams from posts and their associated comments, regardless of stance. We grouped content published through 2024 into one bin and divided content from 2025 through January 2026 into bins with equal numbers of items. This approach accounted for the higher volume of recent content, which would have produced highly uneven groups if we had used equal-length time intervals. We then treated each bin as a class in the c-TF-IDF analysis described above. As a result, the phrases identified for each bin were distinctive to that period rather than merely frequent across the full study period.

## Results

This section presents the results for LLM performance, stance detection, engagement, topic modeling, and temporal trends, organized by research question.

**RQ i— What are the overall public stances toward data center projects? Specifically, we ask a) what is the stance distribution, and b) how popular are the different stances?**

*Stance Detection Performance*

All four candidate LLMs agreed with the human quality-control labels at levels well above chance, with DSv3.1 performing best (Figure 4). Across the two reviewers, DSv3.1 achieved an average Cohen's κ above 0.6, indicating substantial agreement, and approached the inter-reviewer agreement of κ = 0.786. These findings suggest that, when guided by domain-specific labeling rules, an LLM can classify stance with consistency approaching that of human reviewers. We therefore used DSv3.1 to label the full corpus for subsequent analyses.

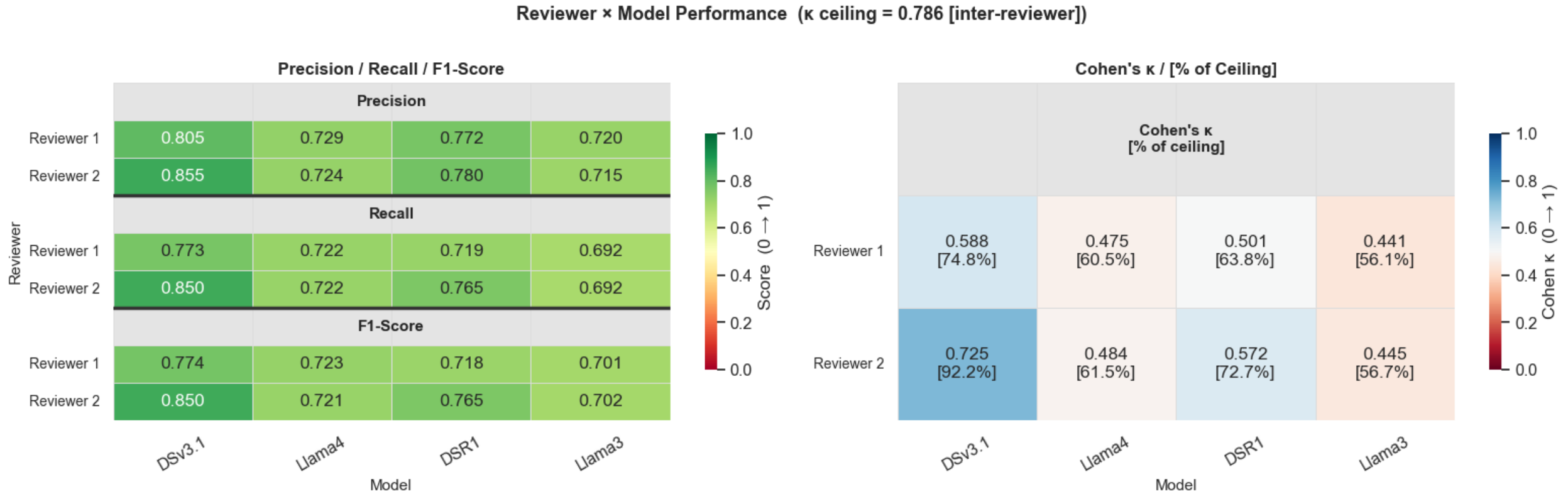

**Figure 4. Precision, recall, F1-score (left panel), and Cohen's Kappa (right panel)**

*Stance Distribution*

Of the 798 filtered posts, stance detection classified 246 (31%) as against, 96 (12%) as favor, and 456 (57%) as none. Although against posts represented less than one-third of all posts, they generated 43,946 comments, which is about 8.5 times more than the 5,154 comments on favor posts (Table 2). Overall, the findings suggest that opposition to data centers generated substantially more public discussion than support did.

**Table 2. Stance distribution results**

| Parameter | Value |
|---|---|
| Number of posts | 795 |
| Number of against posts | 243 |
| Number of favor posts | 96 |
| Number of posts without a stance | 456 |
| **Number of comments associated with against posts** | 43,946 |
| **Number of comments associated with favor posts** | 5,154 |

*Engagement with Competing Positions (RQ i.b)*

We further quantified engagement with the posts using three metrics: upvote score, number of comments, and upvote ratio, as shown in Figure 5.

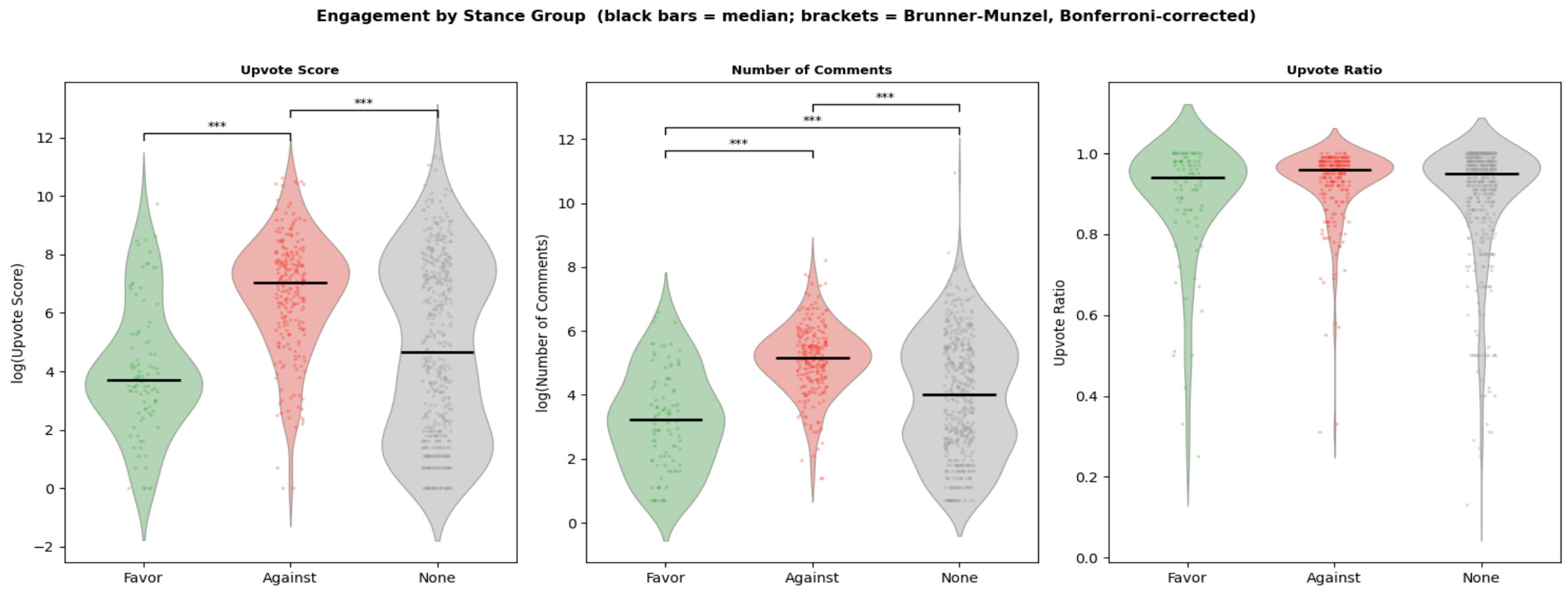

**Figure 5. Reddit user engagement comparison with posts of various stance groups**

Engagement differed significantly by stance. Specifically, against posts attracted significantly higher upvote scores than favor posts (Brunner–Munzel W = 11.24, adjusted $p < .001$, Cliff's $\delta = -0.62$, large effect), as well as than none posts ($\delta = +0.35$, medium effect). As mentioned previously, against posts also generates significantly more comments than both favor ($\delta = -0.66$, large) and none ($\delta = +0.37$, medium) posts. In contrast, the upvote ratio did not differ significantly across stance categories, suggesting that community upvote rates were broadly similar regardless of stance, even though engagement volume varied substantially. Against-oriented content received greater visibility and more overall voting and commenting activity, but "favor content was not disproportionately downvoted. Thus, the findings indicate greater attention to and mobilization around against stance—not necessarily greater hostility toward favor views.

**RQ ii— How do public stances and discourse topics change over time?**

The divide in opinion is longstanding and widening. Weekly volumes of against posts and comments grew faster than favor volumes, and the net difference between against and favor volume widened over the 86-week analysis window (Figure 6). ADF and KPSS tests agreed that every series — including both net-difference series — is non-stationary, indicating that the widening gap reflects a persistent trend rather than random fluctuation.

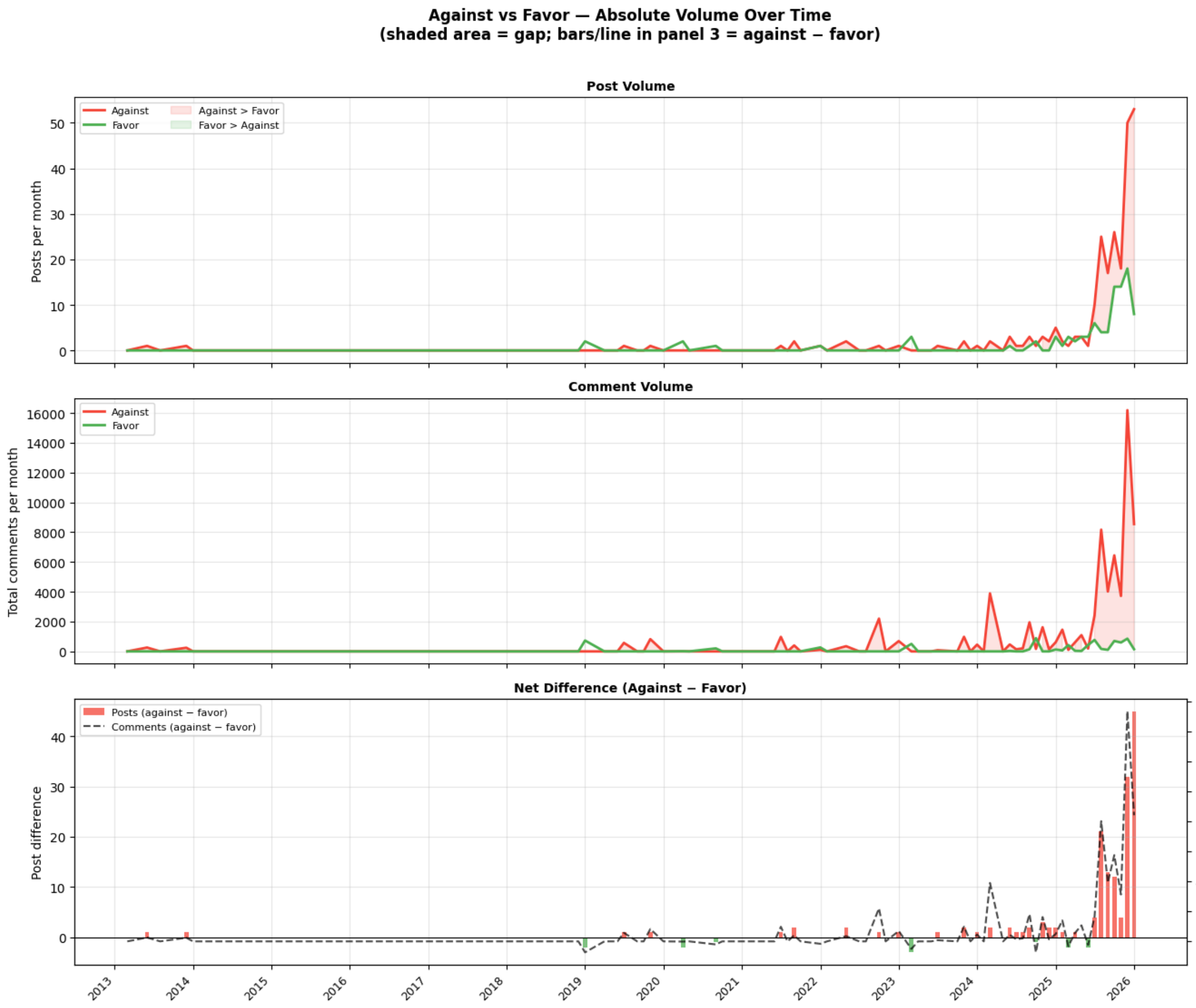

**Figure 6. Time series of post and comments volume by stance groups, and difference in favor and against posts and comments volume over time.**

Table 3 shows how key phrases changed over time. Before 2024, environmental discussion included phrases such as "climate change," "solar panels," and "nuclear power." Phrases specifically about water use appeared in 2025. Energy and water concerns then appeared across successive time bins, such as “water usage”, “water cooling”, “nuclear power”, “closed loop". Because each bin contains the same number of posts and comments but covers a shorter period as time progresses, these findings suggest that environmental discussion became more frequent over time.

**Table 3. c-TF-IDF extracted key phrases over time**

| Time bin | # Posts | # Comments | # Combined | Top phrases |
|---|---|---|---|---|
| **≤2024** | 151 | 28,949 | 29,100 | climate change · years ago · social media · makes sense · long time · solar panels · nuclear power · dc travel |
| **Jan 2025– Aug 2025** | 115 | 18,392 | 18,507 | closed loop · years ago · evaporative cooling · sorry loss · long term · nuclear power · join community · water cooling |
| **Aug 2025– Nov 2025** | 141 | 18,365 | 18,506 | years ago · long term · closed loop · makes sense · water usage · nuclear power · natural gas · power plants |
| **Nov 2025– Dec 2025** | 148 | 18,358 | 18,506 | closed loop · long term · years ago · solar panels · makes sense · power plants · great lakes · evaporative cooling |
| **Dec 2025– Jan 2026** | 243 | 18,264 | 18,507 | long term · years ago · closed loop · tech companies · power plants · big tech · solar panels · water usage |

**RQ iii— What factors shape supportive, opposing, or neutral stances and which topics drive engagement?**

*Reasons Behind Opposition*

Clustering of *against* posts (n = 243) resulted in four argument clusters (Figure 7). Table 4 summarizes the main themes in posts against data center development, along with representative keywords and posts. The primary cluster centered on resource consumption and scarcity (C0; n = 104, 42.8%), framing water and electricity as finite resources consumed at scale and unfairly allocated at community expense. For example, one representative post likened an 800 MW data center campus to adding a “new city” of energy demand, expressing concern that it would use nearly five times as much power as all local homes, businesses, and municipal facilities combined.

The second largest cluster, community resistance and secrecy (C2; n = 83, 34.2%),

highlighted contested approval processes, grid strain, and declining trust among residents, local governments, and developers. For instance, one post questioned why a provincial government would seek to attract a 400 MW data center while warning residents that they were "on the brink of rolling blackouts."

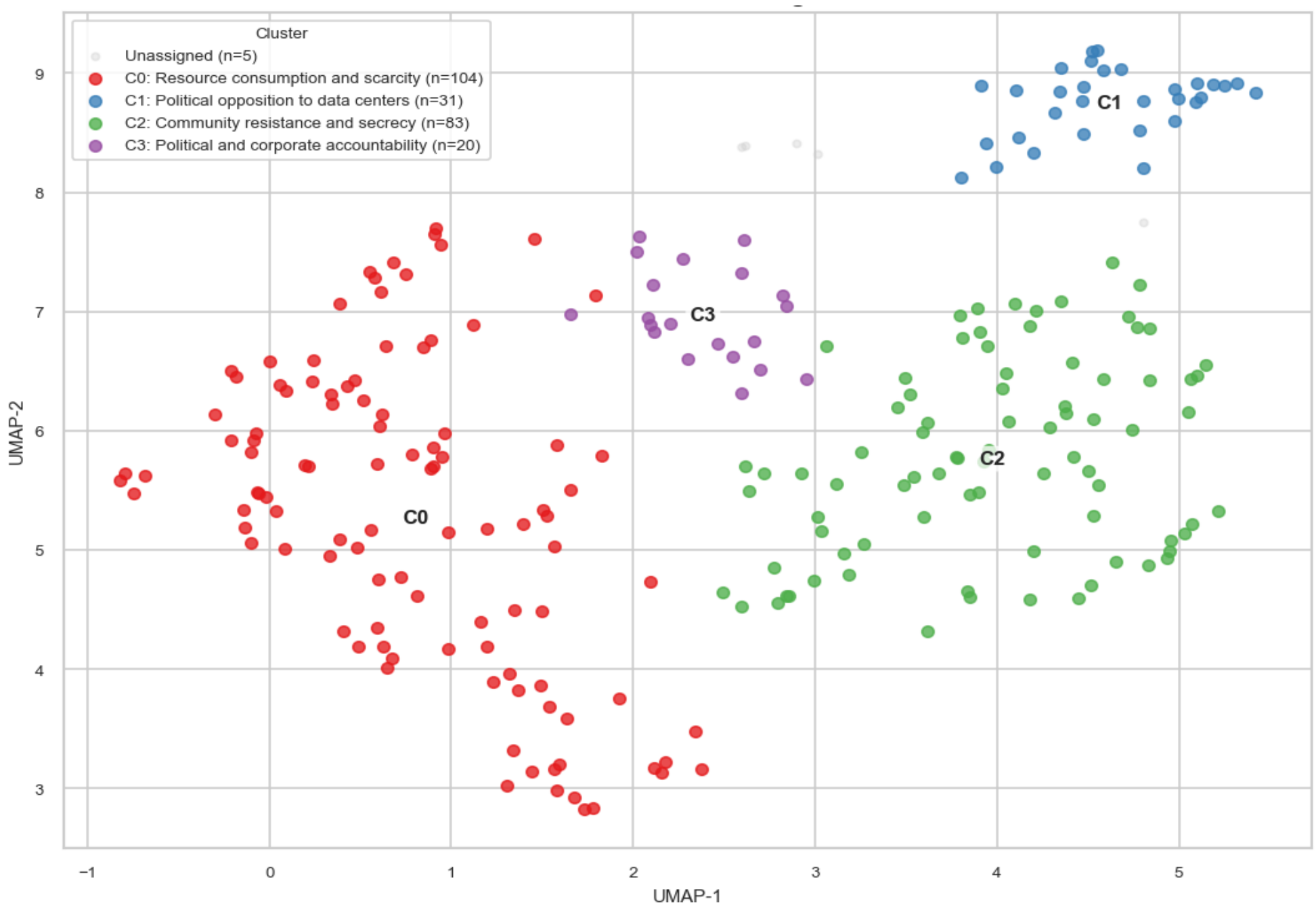

**Figure 7. Clustering results of against stance posts in a reduced 2-dimension space**

The third theme was political opposition to data center development (C1; n = 31, 12.8%). Posts in this theme called for moratoriums or pauses in construction to "give democracy a chance to catch up." The fourth theme was political and corporate accountability (C3; 20 posts, 8.2%). These posts questioned who benefits from data center development and who bears its costs, which one post described "a lot of the costs of operating the data center will be pushed onto the residents." Five posts (2.1%) fell below the density threshold and were left unassigned.

Cluster vocabulary was largely shared rather than distinctive. For instance, C0 shared four of ten terms with C2 and four with C3, and *electricity* ranked in the top terms of three of the four clusters. The non-overlapping remainders were narrow — resource magnitude and fuel (*million*, *gallons*, *gas*) in C0, municipal process and siting (*city*, *planning*, *tax*, *land*) in C2, and proximity and cost incidence (*near*, *proposed*, *built*, *costs*) in C3. C1 was the exception, sharing none of its ten top terms with any other cluster and naming specific actors and actions (*bernie*, *sanders*, *pause*, *construction*), consistent with its clear separation in the projection. Therefore, against posts drew on a common vocabulary of power and water while differing in the frame applied to it, and the clusters separated on that framing rather than on word use. Further sub-clustering analysis on the against posts were performed and results are summarized in Supporting Data (SD) 1.

**Table 4. Summary of clustering results of Reddit posts with cluster topic title, top terms, and representative quotes, presented in descending order of share percentage. Certain parts of the posts were abbreviated, and the full quotes are available as SD 2.**

| Cluster ID | Share % | Topic Title | Top Terms | Representative Quote |
|---|---|---|---|---|
| C0 | 44% | **Resource consumption and scarcity** | water, power, electricity, million, energy, grid, just, new, gallons, gas | … They are planning to resurrect the old Blue Valley Power Plant site to supply the massive power demand… |
| | | | | Phase 1 of construction would construct a 15-turbine, natural-gas-fueled power station capable of generating 225 megawatts... The electricity demand of one 800 MW data center in Independence would dwarf the city’s annual consumption, using nearly five times the power that all homes, businesses, and municipal facilities combined currently use. It's like adding an entire “new city” of energy demand. |
| C2 | 35% | **Community resistance and secrecy** | power, electricity, city, grid, new, planning, tax, demand, fuck, land | NB warns of looming power shortages while planning a massive, power-hungry data center The New Brunswick Government tells us we’re at risk of running out of power. They have also been planning a massive, power-hungry data center in Lorneville since 2024. |
| | | | | Just two nights ago, NB Power Vice-President Brad Coady warned Tantramar councilors that the proposed RIGS gas/diesel plant must be built to ensure “energy security” for New Brunswickers, stating that New Brunswick is at real risk of running out of electricity… |
| | | | | At the same time, provincial agencies, … are actively encouraging a 400-megawatt data-center project in Lorneville… |
| | | | | How does the government reconcile encouraging |
| | | | | development of a data center … while simultaneously telling New Brunswickers that we are on the brink of rolling blackouts … |
| C1 | 13% | **Political opposition to data centers** | bernie, sanders, construction, says, calls, ohio, alberta, microsoft, pause, big | Bernie Sanders wants to temporarily halt AI data center construction nationwide \| He says a pause would "give democracy a chance to catch up" with the rapid buildout. |
| C3 | 8% | **Political and corporate accountability** | water, just, massive, proposed, louisville, near, built, costs, electricity, energy | As much as these articles talk about the influx of jobs being brought in to the community, it is my understanding that a lot of the costs of operating the data center will be pushed onto the residents for a set amount of years before the companies say that they will be able to save money for the community… |

*Reasons Behind Support*

Favor posts were clustered manually due to their small sample size, the results are shown in Table 5. Considering the favor post data quality, only 12 posts were retained for thematic grouping and other posts were not included due to lack of thematic representation or were deemed inaccurately categorized as *favor*. *Economic and investment benefits* has five related posts, spanning state-level investment announcements and retail-investor commentary on data center and semiconductor equities. *Renewable energy integration* has three related posts consisting of corporate procurement and siting announcements, including a facility powered by 1.6 million solar panels and an underwater installation framed around electricity and land savings. *Waste-heat reuse and community benefit* has four related posts, describing heat recovery at domestic and municipal scale, including a shed-housed cluster used for home heating, a public swimming pool in Devon, and Finnish district-heating networks.

Across the three themes, the benefits invoked sat at either macro or micro scale — large scale economic investments on one hand, single-household and single-facility heat recovery on the other.  However, this pattern requires further validation given the relatively small sample size.

**Table 5. Manually assigned themes and post titles for favor-stance posts.**

| # | Theme | Post Title |
|---|---|---|
| 1 | Economic & Investment Benefits | Whitmer supports Michigan's biggest investment ever: Multi-billion dollar AI data center |
| | | Data centers generate 50× more tax revenue per gallon of water than golf courses in Arizona |
| | | $APLD DD AI Datacenters huge run from $3 → $20 since April |
| | | Micron's $100B Megafab Signals Explosive AI Growth — Perfect Tailwind for POET's High-Speed Photonics |
| | | Why Hyperscale data (GPUs) is undervalued and why smart investors are watching |
| 2 | Renewable Energy Integration | Google's new US data centers will run on 1.6 million solar panels as part of its plan to purchase 100% carbon-free energy |
| | | China begins building underwater data center with performance equal to 6 million PCs — aims to save 122 million KWh of electricity and nearly ten soccer fields of land |
| | | Google signs up for $2bn wind and solar investment; becomes the largest corporate buyer of renewable electricity |
| 3 | Waste-Heat Reuse & Community Benefit ommunity Benefit | "I heat my home with a data centre (full of Raspberry Pis) in the shed" |
| | | Essex couple put data centre in their garden shed, save 89% on energy bills |
| | | Heat from a washing-machine-sized data centre used to heat a Devon public swimming pool, saving Exmouth Leisure Centre thousands of pounds |
| | | Finland has started using data centres' waste heat to warm municipal district-heating networks |

*Which Topics Drive Engagement*

Table 6 reports the five highest-engagement bigrams for *against* posts and their associated comments. Post engagement concentrated on a small number of news events: the five leading phrases traced to only three stories — a senator's call to pause new data center construction (*bernie sanders*, *sanders calls*), a Wyoming facility projected to draw five times the power of the state's residents (*human occupants*), and an executive's doubt that AI infrastructure spending would pay off (*ceo says*, *infrastructure costs*).

**Table 6. Top 5 key phrases from against posts and associated comments**

| Top Keywords | Keywords origin | Engagement Metrics | Post/comment quote |
|---|---|---|---|
| bernie sanders | Against posts | engagement=59946 n_posts=7 | Bernie Sanders calls for a pause on new AI data centers in the US. |
| sanders calls | Against posts | engagement=41998 n_posts=3 | |
| human occupants | Against posts | engagement=41641 n_posts=4 | A massive Wyoming data center will soon use 5x more power than the state's human occupants - but no one knows who is using it. |
| ceo says | Against posts | engagement=37534 n_posts=4 | IBM CEO says there is 'no way' spending trillions on AI data centers will pay off at today's infrastructure |
| infrastructure costs | Against posts | engagement36747 n_posts=3 | costs. |
| power plant | Associated comments to against posts | engagement=8114 n_comments=414 | Build your own energy infra then and make sure it doesn't use any water or belch out CO2. If you can't do that, what you're asking for is subsidies to the tune of 30 nuclear power plants. |
| nuclear power | Associated comments to against posts | engagement=6544 n_comments=255 | |
| closed loop | Associated comments to against posts | engagement=6322 n_comments=382 | Evaporative cooling is the most environmentally friendly way of removing heat. We use closed loop cooling only where it's not practical or feasible to employ evaporative cooling. |
| years ago | Associated comments to against posts | Engagement=4948 n_comments=296 | The US used to have a growing renewable energy market and a growing consciousness of how we are hurting our planet. Now we can't opt out of AI while searching online, purchasing online or even are free of it at work. We might have been able to address anthropogenic climate change 20 years ago but now we are accelerating our decline and squeezing the small people for what little they have left. |
| evaporative cooling | Associated comments to against posts | engagement=4174 n_comments=242 | Evaporative cooling is the most environmentally friendly way of removing heat. We use closed loop cooling only where it's not practical or feasible to employ evaporative cooling. |

Comment engagement focused on technical and resource concerns. Four of the five leading phrases referred to energy or cooling infrastructure: *power plant*, *nuclear power*, *closed loop*, and *evaporative cooling*, the last two phrases came from a single thread explaining why data centers consume fresh water. Commenters generally used *nuclear power* to convey the scale of energy demand rather than a preference for nuclear energy. For example, one commenter challenged critics to build the necessary infrastructure without "subsidies to the tune of 30 nuclear power plants." The only nontechnical phrase among the five, *years ago*, appeared in discussions of climate change over time.

Surprisingly, top engaging bigrams from the against posts did not emerge from the largest cluster, as we cross-compare top-engagement phrases against top keywords (unigrams) from clustered posts (Table 6 and Table 4). The highest-engagement post phrases came from C1, the second smallest cluster with only 31 posts (12.8%), while C0, the largest with 104 posts (42.8%), did not contribute any single word that overlapped with top-engagement bigrams from against posts. The comment level on the contrary, showed that discussion was dominated by the energy and water infrastructure keywords that defined C0 — power generation (*power plant*, *nuclear power*) and cooling method (*closed loop*, *evaporative cooling*, both from a thread on data center water consumption) — rather than by the political vocabulary that drove post engagement.

This indicates a clear divide between what drives initial post engagement and what fuels actual community discussion. Peak attention was attributable to a small, news-driven political cluster, while the largest resource-scarcity cluster explains why attention was given to political movement against data center development. Interestingly, leading post phrases accumulated 36.7k–59.9k engagement across just 3–7 posts, whereas leading comment phrases accumulated 4.2k–8.1k across 242–414 comments. This divergence demonstrates that discussion volume and engagement are not interchangeable measures of public concern: ranked by post volume, resource scarcity was the dominant opposition argument, but ranked by engagement, it was nearly absent at the post level. Consequently, focusing solely on top-ranked posts or trending threads over-represents political framing while under-representing the core argument held by the plurality of the opposing population. To put this into perspective, calling for a moratorium on data center developments—the most popular post theme—represents the community's desired outcome, whereas the reasons behind this desire rest within other posts and comment threads.

## Discussion

This study offers two findings with direct implications for how data center development is planned, regulated, and communicated: (1) the distribution of community stances identified through LLM classification and (2) four reproducible clusters of arguments against development. The largest cluster, resource consumption and scarcity, is the focus of several policy initiatives now underway. We discuss each finding in turn and consider the measures it may support, including disclosure requirements, cost-allocation rules, consultation periods, and public education.

**What Communities Ask for Versus What Policy Addresses**

Communities across the United States are rejecting data center projects, this is supported by Data Center Watch (2025) report as well as our findings where several local resistances (e.g., Wyoming, New Brunswickers) express competition between data centers and residents for resources. This resistance points to a gap, and an opportunity, in three distinct classes of policy: (1) disclosure of resource consumption, governing what data centers must report; (2) cost allocation, governing who pays for the infrastructure a facility requires and; (3) process, governing how siting decisions are made and consulted on. Current policies have begun to address the first two issues but have yet to address the third. Our second largest cluster of public comments, community resistance and secrecy (34.2%), highlights the importance of this unaddressed concern.

First, *Disclosure of resource consumption* is the area where existing legislation and our findings most clearly align, although current reporting requirements remain narrow. In our study, community concerns are centered on both water and energy use. Yet facility-level water withdrawal and consumption are not reported reliably or consistently (Carter et al. 2026). Existing mandates are only state-level and resource-specific. For instance, California and Iowa have enacted their own reporting requirements (California State Legislature 2026; Iowa State Legislature 2026) and Michigan has a bill pending (Michigan State Senate 2025). Federal level has introduced measures addressing data center resource use, including the Advancing Water Reuse Act and the Unleashing Low-Cost Rural AI Act, but not enacted (119th U.S. Congress 2025a, b). Our results support disclosure written to cover water and power jointly at the facility level, paired with withdrawal limits and drought contingency terms, because separate water and energy reports may not fully address the combined resource concerns expressed in the largest cluster.

Second, in our study, there are 8.2% of the public opinions discussed around political and corporate accountability, its distinguishing terms—*near*, *proposed*, *built*, and *costs*—pointed to concern about who pays for infrastructure serving data centers. One post described these costs as being "pushed onto the residents". However, *cost allocation* is often addressed through individual utility rate proceedings. For example, Duke Energy Carolinas' 2025–2026 general rate case illustrates this approach: intervenors challenged the allocation of large-load transmission and generation costs across customer classes, and a proposed settlement committed the parties to file a large-load tariff, subject to a final order (North Carolina Utilities Commission 2025). This case shows how a rate proceeding can bring cost allocation into regulatory review, but its outcome depends on the proceeding and the parties involved. Our findings suggested that more consistent rules for assessing and allocating data center infrastructure costs are needed before they are passed on to other customers.

Third, *process transparency* emerges as a policy gap in our findings. Community resistance and secrecy is the second-largest opposition cluster (34.2%). Its distinguishing terms—*city*, *planning*, *tax*, and *land*—suggest that objections concerned how projects were proposed and approved, alongside their physical impacts. One post questioned how a provincial government could pursue a 400 MW facility while warning residents that they were "on the brink of rolling blackouts". Permitting timelines, disclosure of incentive agreements, and consultation requirements vary across jurisdictions and projects. Prior research suggests that opposition may intensify when communities receive information late or distrust the protections offered (Nwebonyi et al. 2022). Publishing incentive agreements before final approval, establishing a consultation period within the permitting schedule, and limiting confidentiality provisions for terms affecting local resource use or tax abatements could help address the concerns about secrecy expressed in these posts.

Lastly, our findings suggest a gap that cannot be addressed by policy alone. *Closed-loop* and *evaporative cooling* were among the comment-level phrases with the highest engagement, drawing 6,322 and 4,174 engagement points across 382 and 242 comments, respectively. Both phrases originated in a single explanatory thread about why data centers consume water. Cooling design involves a tradeoff between water and energy use: evaporative systems consume water to reduce cooling energy demand, whereas closed-loop and air-cooled designs reduce or eliminate water withdrawal at a potential energy-efficiency cost (Kong et al. 2024; Lei et al. 2025; Spenrath 2025). This tradeoff extends beyond data centers because water and energy use are coupled throughout the power system (Rao et al. 2026; Siddik et al. 2021). Disclosure should therefore identify a proposed facility's cooling technology alongside its projected water and power use, making the tradeoff easier to assess. Proactive public education about these tradeoffs may also help communities evaluate proposals and address concerns about resource use.

**Private Industry Response and the Process Gap**

Private firms have begun to address resource use and cost allocation through voluntary commitments. In January 2026, Microsoft pledged to seek electricity rates that cover the costs of serving its data centers without passing those costs on to residential customers. It also committed to improving water-use intensity, replenishing water, and reporting water use by data center region (Microsoft 2026). In June 2026, Google committed to water replenishment, the use of air cooling or recycled water in high-risk watersheds, and annual reporting of water use at its data center locations (Google 2026). Microsoft's rate commitment addresses the cost-allocation concern raised in the Duke Energy Carolinas proceeding, while both firms' water commitments relate to the resource concerns represented by C0, the largest opposition cluster (42.8%).

These commitments have limits as responses to community concerns. Because they are voluntary, their terms and implementation remain under company control. Their reporting measures also differ: a regional water total, a location-level water figure, and a fleet-wide efficiency target do not provide the same basis for comparing individual facilities. Moreover, water replenishment does not necessarily reduce withdrawals at the facility where water is used. The cited commitments address resource use and, in Microsoft's case, electricity costs, but do not establish common requirements for publishing incentive agreements, setting consultation periods, or disclosing terms relevant to siting decisions. Thus, they do not fully address the approval-process concerns represented by C2 (34.2%).

Amazon and Apple further illustrate the difficulty of comparing company-reported measures. Amazon reports a global data center water-use effectiveness of 0.12 liters withdrawn per kilowatt-hour of IT load, an efficiency measure that does not disclose the absolute water withdrawal or electricity use of a named facility (Amazon 2026). Apple reports water replenishment across its offices, data centers, and retail stores, rather than as a directly comparable measure for each data center (Apple 2026). Because both water and power appeared among the leading terms in C0, standardized reporting of both quantities for individual facilities would help residents compare proposed projects with operating sites.

Our findings on favor posts suggest a further opportunity for industry action. Waste-heat reuse and community benefits appeared in four of the 12 retained favorable posts, including examples involving home heating, a public swimming pool in Devon, and Finnish district-heating networks; three other posts highlighted renewable energy integration. Prior research identifies waste heat as a recoverable resource with potential community uses (Sovacool et al. 2022; Terenius et al. 2023). These examples illustrate how a data center could provide a tangible local benefit, although the small number of favorable posts does not establish how widely such benefits would influence community support.

## Conclusion

This study provides a workflow for capturing community stance toward data center development from public social media discourse using LLMs and applies it to 798 Reddit posts and 102,328 associated comments spanning more than a decade. We report four principal findings. First, DeepSeek v3.1, an open-source mixture-of-experts model guided by three domain-specific labeling rules, classified stance at 83% of the human inter-reviewer ceiling (mean Cohen's $\kappa$ = 0.656 against a ceiling of 0.786), establishing that open-weight models are viable stance classifiers for infrastructure discourse and that stance monitoring need not depend on proprietary models or on survey budgets. Second, opposition dominated the discourse by attention rather than by disapproval: against posts outnumbered favor posts 2.6:1 and their comment threads 8.5:1, with large effect sizes on both upvotes and comments, yet upvote ratios did not differ across stance groups. Third, the gap between opposition and support widened across the 86-week analysis window, and every series tested was non-stationary. Fourth, opposition resolved into four reproducible argument clusters: resource consumption and scarcity (4%), contested siting processes and secrecy (34%), political opposition to data centers (13%), and political and corporate accountability (8%). Engagement analysis revealed that posts with emotional or provocative political framing attracted the most engagement, while key technical concerns (e.g., water, power) were dispersed across lower-engagement comment threads.

This study makes three contributions. First, it introduces a critical social perspective to data center development by framing these hyperscale infrastructures not merely as technological assets, but as physical developments that directly contend with local residents for vital community resources; in doing so, it advances the literature beyond traditional, survey- and simulation-based models by leveraging authentic, bottom-up public opinion data to capture genuine community sentiment. Second, by systematically analyzing overall public stances, longitudinal trends, and engagement patterns, this research equips urban planners and decision-makers with crucial empirical evidence to anticipate public reactions, mitigate localized friction, and formulate more informed project policies. Third, the study identifies the granular drivers, themes, and concerns underpinning community support and opposition, shedding light on critical gaps in early-stage infrastructure planning where policy frameworks remain ambiguous. Ultimately, these insights provide actionable strategies for decision-makers to address substantive public concerns, enhance communication transparency, allocate municipal resources equitably, and strike a sustainable equilibrium between rapid technological modernization and long-term community well-being.

## Limitations and Future Research

This research has several limitations. First, Reddit users are not representative of affected communities. Because Reddit was the only platform available for bulk data collection, we cannot separate platform-specific patterns from broader public opinion. Second, post-level stance labels reduce diverse comment threads to a single position. Classification was least precise for the smaller favor category, and the support themes in Table 5 are based on only 12 retained posts. They should therefore be treated as possible themes, not estimates of their prevalence. Third, the vocabulary of the four opposition clusters overlapped: C0 shared four of its ten leading terms with C2 and four with C3. The clusters differ more in how arguments are framed than in the words they use. Future work could also model topics at the comment level and use keyword reranking methods to identify terms that better distinguish the categories. Besides, future work could apply the classifier to geographically defined datasets, such as counties or utility service areas, where siting decisions occur. Linking those data to permitting records could show whether online discussion precedes or follows formal opposition. Future studies could also test whether the proposed transparency measures change community stance after they are adopted.

## Data Availability Statement

Some or all data, models, or code that support the findings of this study are available from the corresponding author upon reasonable request.

## Acknowledgement

The authors thank anonymous colleagues at Carollo Engineers and Dr. Margaret Cook of HARC for their feedback on the manuscript.

## Supplemental Material

Statistical analysis data from BERTopic modeling can be found at https://doi.org/10.5281/zenodo.23320826.